# Design of an Optical Character Recognition System for Camera-based Handheld Devices

Ayatullah Faruk Mollah[1], Nabamita Majumder[2], Subhadip Basu[3] and Mita Nasipuri[4]

[1] School of Mobile Computing and Communication,
Jadavpur University, Kolkata, India

[2] Department of Master of Computer Applications,
MCKV Institute of Engineering, Howrah, India

[3, 4] Department of Computer Science and Engineering,
Jadavpur University, Kolkata, India

**Abstract**
This paper presents a complete Optical Character Recognition (OCR) system for camera captured image/graphics embedded textual documents for handheld devices. At first, text regions are extracted and skew corrected. Then, these regions are binarized and segmented into lines and characters. Characters are passed into the recognition module. Experimenting with a set of 100 business card images, captured by cell phone camera, we have achieved a maximum recognition accuracy of 92.74%. Compared to Tesseract, an open source desktop-based powerful OCR engine, present recognition accuracy is worth contributing. Moreover, the developed technique is computationally efficient and consumes low memory so as to be applicable on handheld devices.

## 1. Introduction

Until a few decades ago, research in the field of Optical Character Recognition (OCR) was limited to document images acquired with flatbed desktop scanners. The usability of such systems is limited as they are not portable because of large size of the scanners and the need of a computing system. Moreover, the shot speed of a scanner is slower than that of a digital camera. Recently, with the advancement of processing speed and internal memory of hand-held mobile devices such as high-end cell-phones, Personal Digital Assistants (PDA), smart phones, iPhones, iPods, etc. having built-in digital cameras, a new trend of research has emerged into picture. Researchers have dared to think of running OCR applications on such devices for having real time results. An automatic Business Card Reader (BCR), meant for automatic population of relevant contact information from a business card also known as *carte de visite* or visiting card into the contact book of the devices, is an example of such applications.

However, computing under handheld devices involves a number of challenges. Because of the non-contact nature of digital cameras attached to handheld devices, acquired images very often suffer from skew and perspective distortion. In addition to that, manual involvement in the capturing process, uneven and insufficient illumination, and unavailability of sophisticated focusing system yield poor quality images.

The processing speed and memory size of handheld devices are not yet sufficient enough so as to run desktop based OCR algorithms that are computationally expensive and require high amount of memory. The processing speeds of mobile devices with built-in camera start with as low as few MHz to as high as 624 MHz. The handset 'Nokia 6600' with an in-built VGA camera contains an ARM9 32-bit RISC CPU having a processing speed of 104 MHz [1]. The PDA 'HP iPAQ 210' has Marvell PXA310 type processor that can compute up to 624 MHz [2]. Some mobile devices have dual processors too. For instance, 'Nokia N95 8GB' has a 'Dual ARM-11 332 M Hz processors' [3]. Processing speed of other mobile phones and PDAs are usually in between them. Besides the lower computing speed, these devices provide limited caching. Random Access Memory (RAM) which is frequently referred to as *internal memory* in case of mobile devices is usually 2-128 MB. Among the mobile devices with high amount of RAM, the following may be mentioned. The PDA 'HP iPAQ 210' has a 128 MB SDRAM [2]. The cell-phone 'Nokia N95 8GB' has a 128 MB internal memory [4]. Compared to desktop computers, this much of memory is very less. In addition to the above challenges, the mobile devices do not have a Floating Point Unit (FPU) [5] which is required for floating point arithmetic operations. However, floating point operations can be performed on such devices by using floating point emulators that result in slower operation. Some more challenges have been reported in [6]. Therefore, need is immensely felt to design computationally efficient and light-weight OCR algorithms for handheld mobile devices.

A number of research works on mobile OCR systems have been found. Laine et al. [7] developed a system for only English capital letters. At first, the captured image is skew corrected by looking for a line having the highest number of consecutive white pixels and by maximizing the given alignment criterion. Then, the image is segmented based on X-Y Tree decomposition and recognized by measuring Manhattan distance based similarity for a set of centroid to boundary features. However, this work addresses only the English capital letters and the accuracy obtained is not satisfactory for real life applications.

Luo et al. of Motorola China Research Center have presented camera based mobile OCR systems for camera phones in [8]-[9]. In [8], a business card image is first down sampled to estimate the skew angle. Then the text regions are skew corrected by that angle and binarized thereafter. Such text regions are segmented into lines and characters, and subsequently passed to an OCR engine for recognition. The OCR engine is designed as a two layer template based classifier. A similar system is presented for Chinese-English mixed script business card images in [9]. In [10], Koga et al. has presented an outline of a prototype Kanji OCR for recognizing machine printed Japanese texts and translating them into English. Another work on Chinese script recognition for business card images is reported in [11]. Moreover, research in developing OCR systems for mobile devices is not limited to document images only. Shen at el. [12] worked on reading LCD/LED displays with a camera phone. A character recognition system for Chinese scripts has been presented in [13].

These studies reflect the feasibility and make a strong indication that OCR systems can be designed for handheld devices. But, of course, the algorithms deployed in these systems must be computation friendly. They should be computationally efficient and low memory consuming.

Under the current work, a character recognition system is presented for recognizing English characters extracted from camera captured image/graphics embedded text documents such as business card images.

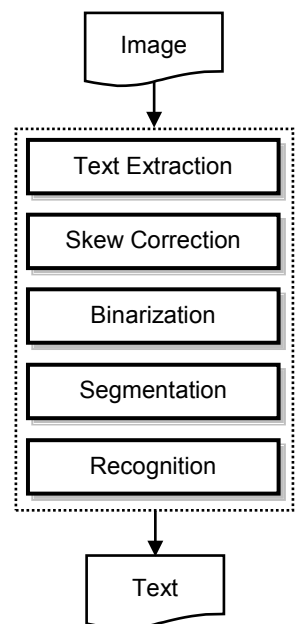

**Fig. 1:** Block diagram of the present system

## 2. Present Work

Modern day handheld devices are usually capable of capturing color images. A color image consists of color pixels represented by a combination of three basic color components viz. red ($r$), green ($g$) and blue ($b$). The range of values for all these color components is 0-255. So, the corresponding gray scale value $f(x,y)$ for each pixel, which also lies between 0-255, may be obtained by using Eq. 1.

$$f(x,y) = 0.299 \times r(x,y) + 0.587 \times g(x,y) + 0.114 \times b(x,y) \quad (1)$$

Applying this transformation for all pixels, the gray scale image is obtained and is represented as a matrix of gray level intensities, $I_{PxQ} = [f(x,y)]_{PxQ}$ where $P$ and $Q$ denote the number of rows i.e. the height of the image and the number of the columns i.e. the width of the image respectively. $f(x,y) \in G_L = \{0,1,\ldots,L-1\}$, the set of all gray levels, where $L$ is the total number of gray levels in the image. Such a gray level image is fed as input to the proposed character recognition system.

The block diagram of the present character recognition system is shown in Fig. 1. The input gray level image is segmented into two types of regions – Text Regions (TR) and Non-text Regions (NR) as illustrated in Sec. 2.1. The NRs are removed and the TRs are de-skewed as discussed in Sec 2.2. In Sec. 2.3 and 2.4, the de-skewed TRs are binarized and segmented into lines and characters. Finally, the characters are recognized as illustrated in Sec. 2.5.

## 2.1 Text Region Extraction

The input image $I_{PxQ}$ is, at first, partitioned into $m$ number of blocks $B_i$, i=1,2,…,$m$ such that $B_i \cap B_j = \emptyset$ and $I_{PxQ} = \bigcup_{i=1}^{m} B_i$. A block $B_i$ is a set of pixels represented as $B_i = [f(x,y)]_{HxW}$ where $H$ and $W$ are the height and the width of the block respectively. Each individual $B_i$ is classified as either Information Block (IB) or Background Block (BB) based on the intensity variation within it. After removal of BBs, adjacent/contiguous IBs constitute isolated components called as regions, $R_i$, i=1,2, ..,$n$ such that $R_i \cap R_j = \emptyset$ for all $i \neq j$ but $\bigcup_{i=1}^{n} R_i \neq I_{PxQ}$ because some BBs have got removed. The area of a region is always a multiple of the area of the blocks. These regions are then classified as TR or NR using various characteristics features of textual and non-textual regions such as dimensions, aspect ratio, information pixel density, region area, coverage ratio, histogram, etc. A detail description of the technique has been reported in one of our prior work [14]. Fig. 2 shows a camera captured image and the TRs extracted from it.

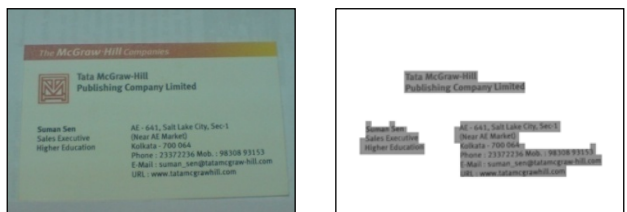

Fig. 2. A camera captured image and the text regions extracted from it

## 2.2 Skew Correction

Camera captured images very often suffer from skew and perspective distortion as discussed in Section 1. These occur due to unparallel axes and/or planes at the time of capturing the image. The acquired image does not become uniformly skewed mainly due to perspective distortion. Skewness of different portions of the image may vary between $+\alpha$ to $-\beta$ degrees where both $\alpha$ and $\beta$ are +ve numbers. Therefore, the image cannot be de-skewed at a single pass. On the other hand, the effect of perspective distortion is distributed throughout the image. Its effect is hardly visible within a small region (e.g. the area of a character) of the image. At the same time, we see that the image segmentation module generates only a few text regions. So, these text regions are de-skewed using a computationally efficient and fast skew correction technique designed in our work and published in [15]. A brief description has been given here.

Every text region has two types of pixels – dark and gray. The dark pixels constitute the texts and the gray pixels are background around the texts. For the four sides of the virtual bounding rectangle of a text region, we can have four sets of values that will be called as profiles. If the length and breadth of the bounding rectangle are $M$ and $N$ respectively, then two profiles will have $M$ number of values each and the other two will have $N$ number of values each. These values are the distances in terms of pixel from a side to the first gray/black pixel of the text region. Among these four profiles, the one which is from the bottom side of the text region is taken into consideration for estimating skew angle as shown in Fig. 3. This bottom profile is denoted as $\{h_i, i=1,2,…,M\}$.

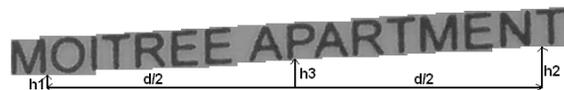

Fig. 3. Calculation of skew angle from bottom profile of a text region

The mean ($\mu = \frac{1}{M}\sum_{i=1}^{M} h_i$) and the first order moment ($\tau = \frac{1}{M}\sum_{i=1}^{M} |\mu - h_i|$) of $h_i$ values are calculated. Then, the profile size is reduced by excluding some $h_i$ values that are not within the range, $\mu \pm \tau$. The central idea behind this exclusion is that these elements hardly contribute to the actual skew of the text region. Now, from the remaining profile elements, we choose the leftmost $h1$, right-most $h2$ and the middle one $h3$. The final skew angle is computed by averaging the three skew angles obtained from the three pairs $h1$-$h3$, $h3$-$h2$ and $h1$-$h2$. Once the skew angle for a text region is estimated, it is rotated by the same.

## 2.3 Binarization

A skew corrected text region is binarized using a simple yet efficient binarization technique [16] developed by us before segmenting it. The algorithm has been given below. Basically, this is an improved version of Bernsen's binarization method [17]. In his method, the arithmetic mean of the maximum ($G_{max}$) and the minimum ($G_{min}$) gray levels around a pixel is taken as the threshold for binarizing the pixel. In the present algorithm, the eight immediate neighbors around the pixel subject to binarization are also taken as deciding factors for binarization. This type of approach is especially useful to connect the disconnected foreground pixels of a character.

```
Binarization algorithm
begin
    for all pixels (x, y) in a TR
        if intensity(x, y) < (G_min + G_max)/2, then
            mark (x, y) as foreground
        else
            if no. of foreground neighbors > 4, then
                mark (x, y) as foreground
            else
                mark (x, y) as background
            end if
        end if
    end for
end
```

### 2.4 Text Region Segmentation

After binarizing a text region, the horizontal histogram profile $\{f_i, i = 1,2,..,H_R\}$ of the region as shown in Fig. 4 is analyzed for segmenting the region into text lines. Here $f_i$ denotes the number of black pixel along the $i^{th}$ row of the TR and $H_R$ denotes the height of the de-skewed TR. At first, all possible line segments are determined by thresholding the profile values. The threshold is chosen so as to allow over-segmentation. Text line boundaries are referred by the values of $i$ for which the value of $f_i$ is less than the threshold. Thus, $n$ such segments represent $n$-1 text lines. After that the inter-segment distances are analyzed and some segments are rejected based on the idea that the distance between two lines in terms of pixels will not be too small and the inter-segment distances are likely to become equal. A detail description of the method is given in [18]. Using vertical histogram profile of each individual text lines, words and characters are segmented. Sample segmented characters have been shown in Fig. 5.

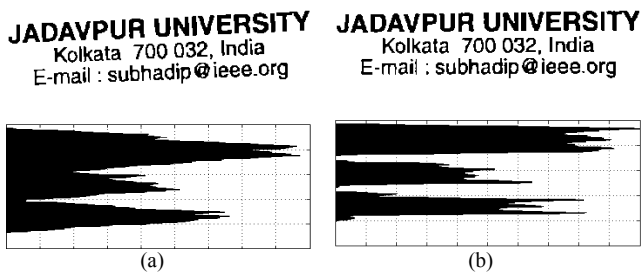

(a) (b)
Fig. 4. Use of horizontal histogram of text regions for their segmentation
   (a) A skewed text region and its horizontal histogram,
   (b) Skew corrected text region and its horizontal histogram

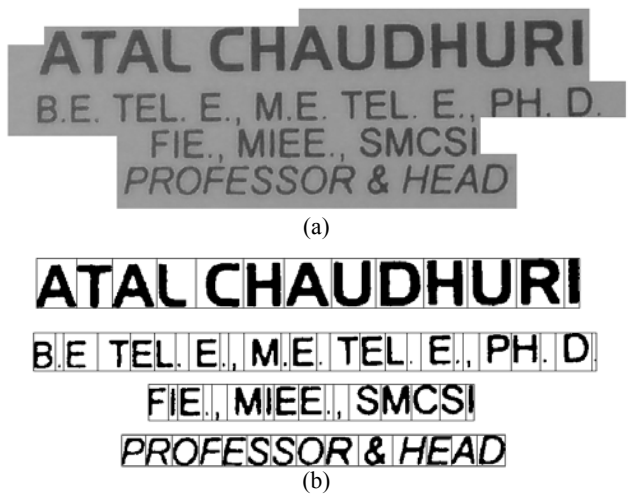

(a)

(b)

Fig. 5. Skew correction and segmentation of text regions,
   (a) An extracted text region,
   (b) Characters segmented from de-skewed text region

### 2.5 Character Recognition

Fig. 6 shows the sequential steps required to classify an individual binarized character. After resizing the pattern by its bounding box, it is normalized to a standard dimension, 48x48.

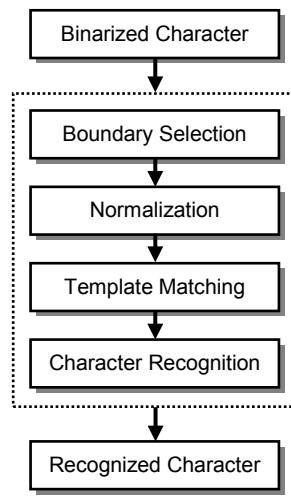

Fig. 6. Block diagram of the character recognition module

Among the 256 ASCII characters, only 94 are used in document images. And among these 94 characters, only 73 are frequently used. In the present scope of experiment, we have considered 73 classes recognition problem. These 73 characters are listed in Table 1. These include 26 capital letters, 26 small letters, 10 numeric digits and 11 special characters.

Table 1: 73 Classes Recognition Problem

| # | & | ( | ) | + | , | - | . | / | 0 |
|---|---|---|---|---|---|---|---|---|---|
| 1 | 2 | 3 | 4 | 5 | 6 | 7 | 8 | 9 | : |
| @ | A | B | C | D | E | F | G | H | I |
| J | K | L | M | N | O | P | Q | R | S |
| T | U | V | W | X | Y | Z | a | b | c |
| d | e | f | g | h | i | j | k | l | m |
| n | o | p | q | r | s | t | u | v | w |
| x | y | z |   |   |   |   |   |   |   |

It may be noted from Table 1 that some classes are similar with some other classes. Such similar classes have been listed in Table 2. Experiments reveal that the classification accuracy gets considerably reduced because of misclassification among these similar classes. Therefore, each such group of similar classes in the 73 class recognition problem will be considered as a single class. In the post-processing phase followed by recognition, these classes can be again identified. Thus, we get a reduced set of 62 classes and it is called as 62 class recognition problem.

Table 2: List of Symmetric Classes

| Symmetric Classes | Merged Class |
|---|---|
| **C**(Capital), **c**(Small) | C |
| **0**(Zero), **O**(Capital), **o**(Small) | O |
| **S**(Capital), **s**(Small) | S |
| **U**(Capital), **u**(Small) | U |
| **V**(Capital), **v**(Small) | V |
| **W**(Capital), **w**(Small) | W |
| **Z**(Capital), **z**(Small) | Z |
| **I**(Capital I), **l** (Small L), **1**(One) | I |

In the present scope of experiments, both 73 class and 62 class recognition problems have been experimented and the results obtained have been reported.

10 representative samples from each class have been taken as the standard templates of that particular class. More samples could have been taken, but that would increase the classification time, a considerable matter of concern in the present work. Experiments show that 10 is a good choice.

A template is represented by a matrix, $T_{NxN} = [t(x,y)]_{NxN}$ where $NxN$ is the template size and $t(x,y) \in \{0,1\}$. Let us denote the test pattern as $T'_{NxN} = [t'(x,y)]_{NxN}$. Now, the correlation ($C$) between a template and a test pattern is calculated as shown in Eq. 2.

$$C = \sum_{x \in N, y \in N} |t(x,y) - t'(x,y)| \quad (2)$$

The class of the template for which the *maximum correlation* is obtained is the class of the test pattern.

## 3. Experimental Results and Discussion

Experiments have been carried out to evaluate the performance and applicability of the current technique for character recognition on a set of 100 business card images of wide varieties. The images have been captured with a cell phone (Sony Ericsson K810i) camera.

As discussed in Sec. 2.1, a segmented region $R_i \in \{TR, NR\}$. Based on the presence of a region $R_i$ in either or both ground truth and output images, it is classified as true positive (*TP*), false positive (*FP*), true negative (*TN*) and false negative (*FN*). So, F-Measure (*FM*) can be defined in terms of recall rate (*R*) and precision rate (*P*) can be given as Eq. 3.

$$FM = \frac{2 \times R \times P}{R + P} \quad (3)$$

where $R = \frac{TP}{TP+FN}$ and $P = \frac{TP}{TP+FP}$. Here, *TP*, *FP*, *TN* and *FN* denote their respective counts. In an ideal situation i.e. when the output image is identical with the ground truth image, *R*, *P* and *FM* should be all 100%. In the present experiment, we have found an F-Measure of 99.48%. This indicates that only a few text regions are improperly segmented. Detail experimentation and results have been given in [14]. Segmented text regions are then skew corrected and binarized. The skew correction technique is very fast and satisfactorily accurate. The skew angle error is between ±3 degrees only.

Although binarization is done after skew correction in the present work, binarization accuracy is calculated without skew correction. To quantify the binarization accuracy, similar method as discussed above has been adopted. Pixel-wise ground truth images are compared with that of the output binarized images. Thus, we get the counts of *TP*, *FP*, *TN* and *FN*. From these counts, we get an average recall rate of 93.52%, precision rate of 96.27% and *FM* of 94.88%.

After binarizing the skew corrected text regions, it is segmented into text lines. The proposed text region segmentation method has been found to be effective enough to successfully segment all text regions in the present scope of experiment. It means that no instance is found where two or more text lines have been extracted as a single text line or a single text line gets treated as more than one text lines. Besides having skew corrected printed texts, soundness of the segmentation method makes text region segmentation so accurate. Character level segmentation accuracy is 97.48% obtained with 3 MP images. It may be noted that the present segmentation technique is not meant for italic and cursive texts. So, such texts have been ignored while calculating the segmentation accuracy.

Recognition accuracy has been shown in Table 3. The maximum recognition accuracy has been achieved with 62 class recognition problem. A close observation reveals that most misclassifications are basically of a few kinds such as between 'B' and '8', ',' and '.', etc.

Table 3: Percentage of recognition with different classifiers

| Classifier | No. of Test Patterns | No. of Correct Classification | % of Recognition |
|---|---|---|---|
| Template Matching (62 Class) | 15807 | 14659 | 92.74% |
| Template Matching (73 Class) | 15807 | 13089 | 82.81% |
| Tesseract [19] | 15807 | 14781 | 93.51% |

The applicability of the present technique on handheld/ mobile devices has been studied in terms of computational requirements. The average time consumption for recognizing a business card image of 3 megapixel resolution is 1.25 seconds with respect to a moderately powerful computer (DualCore T2370, 1.73 GHz, 1GB RAM, 1MB L2 Cache). It may be noted that the technique seems to be fast enough.

## 4. Conclusion

A complete OCR system has been presented in this paper. Because of the computing constraints of handheld devices, we have kept our study limited to light-weight and computationally efficient techniques. Compared to Tesseract, acquired recognition accuracy (92.74%) is good enough. Experiments shows that the recognition system presented in this paper is computationally efficient which makes it applicable for low computing architectures such as mobile phones, personal digital assistants (PDA) etc.

Recognition is often followed by a post-processing stage. We hope and foresee that if post-processing is done, the accuracy will be even higher and then it could be directly implemented on mobile devices. Implementing the presented system with post-processing on mobile devices is also taken as part of our future work.

Acknowledgments

Authors are thankful to the *Center for Microprocessor Application for Training Education and Research* (CMATER) and project on *Storage Retrieval and Understanding of Video for Multimedia* (SRUVM) of the Department of Computer Science and Engineering, Jadavpur University for providing infrastructural support for the research work. We are also thankful to the *School of Mobile Computing and Communication* (SMCC) for providing the research fellowship to the first author. The second author is thankful to MCKV Institute of Engineering for allowing her in carrying out research.

**Ayatullah Faruk Mollah** received his B.E. degree in Computer Science and Engineering from Jadavpur University, Kolkata, India in 2006. Then he served as a Senior Software Engineer in Atrenta (I) Pvt. Ltd., Noida, India for two years. He is now a Ph.D. student in the School of Mobile Computing and Communications of Jadavpur University. His research interests include image processing and pattern recognition on low computing platforms. He is also a student member of IEEE.

**Nabamita Majumder** received her Master of Computer Application (MCA) degree from Indira Gandhi National Open University, Delhi, India in 2006. She received M.Tech degree in Software Engineering from West Bengal University of Technology, Kolkata, India in 2010. She served as a faculty member of DOEACC Center Kolkata for 2 years. She is now Assistant professor in the Department of MCA of MCKV Institute of Technology, West Bengal, India.

**Subhadip Basu** received his B.E. degree in Computer Science and Engineering from Kuvempu University, Karnataka, India, in 1999. He received his Ph.D. (Engg.) degree thereafter from Jadavpur University (J.U.) in 2006. He joined J.U. as a senior lecturer in 2006. His areas of current research interest are OCR of handwritten text, gesture recognition, real-time image processing.

**Mita Nasipuri** received her B.E.Tel.E., M.E.Tel.E., and Ph.D. (Engg.) degrees from Jadavpur University, in 1979, 1981 and 1990, respectively. Prof. Nasipuri has been a faculty member of J.U since 1987. Her current research interest includes image processing, pattern recognition, and multimedia systems. She is a senior member of the IEEE, USA, Fellow of I.E (India) and W.A.S.T., Kolkata, India.